\documentclass[11pt]{article}

\usepackage{acl2005}
\usepackage{times}
\usepackage{latexsym}
\usepackage[dvips]{graphics}
\usepackage{epsfig}
\usepackage{xspace}
\usepackage{relsize}
\usepackage{url}

\title{Bootstrapping Deep Lexical Resources: Resources for Courses}

\author{Timothy Baldwin\\
   Department of Computer Science and Software Engineering\\
   University of Melbourne, Victoria 3010 Australia\\
   \url{tim@csse.unimelb.edu.au}}

\date{}

\newcommand{\example}[1]{\textit{#1}}
\newcommand{\tx}[1]{\textbf{#1}}
\newcommand{\lextype}[1]{\texttt{#1}}

\newcommand{\tabref}[1]{Table~\ref{#1}\xspace}
\newcommand{\figref}[2][]{Figure#1~\ref{#2}\xspace}
\newcommand{\secref}[2][]{Section#1~\ref{#2}\xspace}
\newcommand{\tabheader}[1]{\textit{\bfseries #1}}
\newcommand{\raisedtab}[1]{\raisebox{1.5ex}[0pt]{#1}}
\newcommand{\tablabel}[1]{\textit{#1}}

\newcommand{\ngram}[1][]{$n$-gram#1\xspace}

\newcommand{\highavail}{$^{***}$\xspace}
\newcommand{\medavail}{$^{**}$\xspace}
\newcommand{\lowavail}{$^{*}$\xspace}

\newcommand{\wordclass}[1]{$_\mathsf{#1}$}
\newcommand{\wcnoun}{\wordclass{N}}
\newcommand{\wcverb}{\wordclass{V}}
\newcommand{\wcadj}{\wordclass{Adj}}

\newcommand{\modifier}[1][]{Modifier$_\mathsf{#1}$}
\newcommand{\modifiee}[1][]{Modifiee$_\mathsf{#1}$}
\newcommand{\conj}[1][]{Conj$_\mathsf{#1}$}

\newcommand{\mycaption}[1]{\caption{\setlength{\baselineskip}{\bskipfactor\baselineskip}#1}}

\def\bskipfactor{0.942}

\newcommand{\captionfonts}{\smaller[0]}

\makeatletter  
\long\def\@makecaption#1#2{%
  \vskip\abovecaptionskip
  \sbox\@tempboxa{{\captionfonts #1: #2}}%
  \ifdim \wd\@tempboxa >\hsize
    {\setlength{\baselineskip}{\bskipfactor\baselineskip}\captionfonts #1: #2\par}
  \else
    \hbox to\hsize{\hfil\box\@tempboxa\hfil}%
  \fi
  \vskip\belowcaptionskip}
\makeatother

\begin{document}

\setlength\titlebox{4cm}    
\setlength{\baselineskip}{\bskipfactor\baselineskip}

\maketitle

\begin{abstract}
  We propose a range of deep lexical acquisition methods which make use
  of morphological, syntactic and ontological language resources to
  model word similarity and bootstrap from a seed lexicon. The different
  methods are deployed in learning lexical items for a precision
  grammar, and shown to each have strengths and weaknesses over
  different word classes. A particular focus of this paper is the
  relative accessibility of different language resource types, and
  predicted ``bang for the buck'' associated with each in deep lexical
  acquisition applications.
\end{abstract}

\section{Introduction}
\label{sec:introduction}

Over recent years, computational linguistics has benefitted considerably
from advances in statistical modelling and machine learning, culminating
in methods capable of deeper, more accurate automatic analysis, over a
wider range of languages. Implicit in much of this work, however, has
been the existence of \tx{deep language resources} (DLR hereafter)
of ever-increasing linguistic complexity, including lexical semantic
resources (e.g.\ WordNet and FrameNet), precision grammars (e.g.\ the
English Resource Grammar and the various ParGram grammars) and
richly-annotated treebanks (e.g.\ PropBank and CCGbank).

Due to their linguistic complexity, DLRs are invariably constructed by
hand and thus restricted in size and coverage. Our aim in this paper is
to develop general-purpose automatic methods which can be used to
automatically expand the coverage of an existing DLR, through the
process of \tx{deep lexical acquisition} (DLA hereafter).

The development of DLRs can be broken down into two basic tasks: (1)
design of a data representation to systematically capture the
generalisations and idiosyncracies of the dataset of interest
(\tx{system design}); and (2) classification of data items according to
the predefined data representation (\tx{data classification}). In the
case of a deep grammar, for example, system design encompasses the
construction of the system of lexical types, templates, and/or phrase
structure rules, and data classification corresponds to the
determination of the lexical type(s) each individual lexeme conforms to.
DLA pertains to the second of these tasks, in automatically mapping a
given lexeme onto a pre-existing system of lexical types associated with
a DLR.

We propose to carry out DLA through a bootstrap process, that is by
employing some notion of word similarity, and learning the lexical types
for a novel lexeme through analogy with maximally similar word(s) for
which we know the lexical types. In this, we are interested in exploring
the impact of different secondary language resources (LRs) on DLA, and
estimating how successfully we can expect to learn new lexical items
from a range of LR types. That is, we estimate the expected DLA ``bang
for the buck'' from a range of secondary LR types of varying size and
complexity. As part of this, we look at the relative impact of different
LRs on DLA for different open word classes, namely nouns, verbs,
adjectives and adverbs.


We demonstrate the proposed DLA methods relative to the English Resource
Grammar (see \secref{sec:erg}), and in doing so assume the lexical types
of the target DLR to be syntactico-semantic in nature. For example, we
may predict that the word \example{dog} has a usage as an intransitive
countable noun (\lextype{n\_intr\_le},\footnote{All example lexical
  types given in this paper are taken directly from the English Resource
  Grammar -- see \secref{sec:erg}.} cf.\ \example{The \underline{dog}
  barked}), and also as a transitive verb (\lextype{v\_np\_trans\_le},
cf.\ \example{It \underline{dogged} my every step}).  

A secondary interest of this paper is the consideration of how well we
could expect to perform DLA for languages of differing density, from
``low-density'' languages (such as Walpiri or Uighur) for which we have
limited LRs, to ``high-density'' languages (such as English or Japanese)
for which we have a wide variety of LRs. To this end, while we
exclusively target English in this paper, we experiment with a range of
LRs of varying complexity and type, including morphological, syntactic
and ontological LRs. Note that we attempt to maintain consistency across
the feature sets associated with each, to make evaluation as equitable
as possible.

The remainder of this paper is structured as follows. \secref{sec:task}
outlines the process of DLA and reviews relevant resources and
literature. \secref[s]{sec:morphology}, \ref{sec:corpus} and
\ref{sec:ontology} propose a range of DLA methods based on morphology,
syntax and ontological semantics, respectively.  \secref{sec:evaluation}
evaluates the proposed methods relative to the English Resource Grammar.

\section{Task Outline}
\label{sec:task}

This research aims to develop methods for DLA which can be run
automatically given: (a) a pre-existing DLR which we wish to expand the
coverage of, and (b) a set of secondary LRs/preprocessors for that
language. The basic requirements to achieve this are the discrete
inventory of lexical types in the DLR, and a pre-classification of each
secondary LR (e.g.\ as a corpus or wordnet, to determine what set of
features to employ).  Beyond this, we avoid making any assumptions about
the language family or DLR type.

The DLA strategy we propose in this research is to use secondary LR(s)
to arrive at a feature signature for each lexeme, and map this onto the
system of choice indirectly via supervised learning, i.e.\ observation
of the correlation between the feature signature and classification of
bootstrap data. This methodology can be applied to unannotated corpus
data, for example, making it possible to tune a lexicon to a particular
domain or register as exemplified in a particular repository of text. As
it does not make any assumptions about the nature of the system of
lexical types, we can apply it fully automatically to any DLR and feed
the output directly into the lexicon without manual intervention or
worry of misalignment. This is a distinct advantage when the inventory
of lexical types is continually undergoing refinement, as is the case
with the English Resource Grammar (see below).

A key point of interest in this paper is the investigation of the
relative ``bang for the buck'' when different types of LR are used for
DLA. 
Crucially, we investigate only LRs which we believe to be plausibly
available for languages of varying density, and aim to minimise
assumptions as to the pre-existence of particular preprocessing tools.
The basic types of resources and tools we experiment with in this paper
are detailed in \tabref{tab:resource-types}.

\begin{table*}[t]
  
  \centering
  \begin{tabular}{lll}
    \hline
    \\
    \raisedtab{\tabheader{Secondary LR type}} & \raisedtab{\tabheader{Description}} & \raisedtab{\tabheader{Preprocessor(s)}}\\
    \hline

    Word list\highavail & List of words with basic POS & --- \\ 
    
    Morphological lexicon\lowavail & Derivational and inflectional
    word relations & --- \\
    
    Compiled corpus\highavail & Unannotated text corpus & POS tagger\medavail \\
    && Chunk parser\lowavail \\
    && Dependency parser\lowavail \\

    WordNet-style ontology\lowavail & Lexical semantic word linkages & --- \\
  \end{tabular}
  \mycaption{Secondary LR and tool types targeted in this research (\highavail
    = high expectation of availability for a given language; \medavail = medium expectation
    of availability; \lowavail = low expectation
    of availability)}
  \label{tab:resource-types}
\end{table*}

Past research on DLA falls into two basic categories: expert
system-style DLA customised to learning particular linguistic
properties, and DLA via resource translation.  In the first instance, a
specialised methodology is proposed to (automatically) learn a particular
linguistic property such as verb subcategorisation (e.g.\
\newcite{Korhonen:2002}) 
or noun countability (e.g.\ \newcite{Baldwin:Bond:2003a}), and little
consideration is given to the applicability of that method to more
general linguistic properties. In the second instance, we take one DLR
and map it onto another to arrive at the lexical information in the desired
format. This can take the form of a one-step process, in mining lexical
items directly from a DLR (e.g.\ a machine-readable dictionary
\cite{Sanfilippo:Poznanski:1992})
, or two-step process in reusing an existing system to learn lexical
properties in one format and then mapping this onto the DLR of choice
(e.g.\ \newcite{Carroll:Fang:2004} for verb subcategorisation learning).

There have also been instances of more general methods for DLA, aligned
more closely with this research.  \newcite{Fouvry:2003b} proposed a
method of token-based DLA for unification-based precision grammars,
whereby partially-specified lexical features generated via the
constraints of syntactically-interacting words in a given sentence
context, are combined to form a consolidated lexical entry for that
word.  That is, rather than relying on indirect feature signatures to
perform lexical acquisition, the DLR itself drives the incremental
learning process.  Also somewhat related to this research is the
general-purpose verb feature set proposed by \newcite{Joanis:2003},
which is shown to be applicable in a range of DLA tasks relating to
English verbs.

\subsection{English Resource Grammar}
\label{sec:erg}

All experiments in this paper are targeted at the \tx{English Resource
  Grammar} (ERG; \newcite{Flickinger:2002}, \newcite{Copestake:Flickinger:2000}).
The ERG is an implemented open-source broad-coverage precision
Head-driven Phrase Structure Grammar (HPSG) 
developed for both parsing and generation.  It contains roughly 10,500
lexical items, which, when combined with 59 lexical rules, compile out
to around 20,500 distinct word forms.\footnote{All statistics and
  analysis relating to the ERG in this paper are based on the version of
  11 June, 2004.}  Each lexical item consists of a unique identifier, a
lexical type (one of roughly 600 leaf types organized into a type
hierarchy with a total of around 4,000 types), an orthography, and a
semantic relation. The grammar also contains 77 phrase structure rules
which serve to combine words and phrases into larger constituents. 
Of the 10,500 lexical items, roughly 3,000 are multiword expressions.

To get a basic sense of the syntactico-semantic granularity of the ERG,
the noun hierarchy, for example, is essentially a cross-classification of
countability/determiner co-occurrence, noun valence and preposition
selection properties. For example, lexical entries of
\lextype{n\_mass\_count\_ppof\_le} type can be either countable or
uncountable, and optionally select for a PP headed by \example{of}
(example lexical items are \example{choice} and
\example{administration}).

As our target lexical type inventory for DLA, we identified all
open-class lexical types with at least 10 lexical entries, under the
assumption that: (a) the ERG has near-complete coverage of closed-class
lexical entries, and (b) the bulk of new lexical entries will correspond
to higher-frequency lexical types. This resulted in the following
breakdown:\footnote{Note that all results are over simplex lexemes only,
  and that we choose to ignore multiword expressions in this research.}
\begin{center}
  \begin{tabular}{@{\hspace{2mm}}lcr@{\hspace{10mm}}}
    \multicolumn{1}{c}{\tabheader{Word class}} & \multicolumn{1}{c}{\tabheader{Lexical types}} & \multicolumn{1}{c}{\tabheader{Lexical items}} \\
    \hline
    Noun & 28 & 3,032\\
    Verb & 39 & 1,334 \\
    Adjective & 17 & 1,448 \\
    Adverb & 26 & 721 \\
    \hline
    Total & 110 & 5,675 \\
  \end{tabular}
\end{center}
Note that it is relatively common for a lexeme to occur with more than
one lexical type in the ERG: 22.6\% of lexemes have more than one
lexical type, and the average number of lexical types per lexeme is 1.12.

In evaluation, we assume we have prior knowledge of the basic word
classes each lexeme belongs to (i.e.\ noun, verb, adjective and/or
adverb), information which could be derived trivially from pre-existing
shallow lexicons and/or the output of a tagger.

Recent development of the ERG has been tightly coupled with treebank
annotation, and all major versions of the grammar are deployed over a
common set of treebank data to help empirically trace the evolution of
the grammar and retrain parse selection models \cite{Oepen+:2002}. We
treat this as a held-out dataset for use in analysis of the \textit{token}
frequency of each lexical item, to complement analysis of
\textit{type}-level learning performance (see \secref{sec:evaluation}).

\subsection{Classifier design}
\label{sec:classifier}

The proposed procedure for DLA is to generate a feature signature for
each word contained in a given secondary LR, take the subset of lexemes
contained in the original DLR as training data, and learn lexical items
for the remainder of the lexemes through supervised learning. In order
to maximise comparability between the results for the different DLRs, we
employ a common classifier design wherever possible (in all cases other
than ontology-based DLA), using TiMBL 5.0 \cite{Daelemans+:2003}; we
used the IB1 $k$-NN learner implementation within TiMBL, with $k$ = 9
throughout.\footnote{We also experimented with bsvm and SVMLight, and a
  maxent toolkit, but found TiMBL to be superior overall, we hypothesise
  due to the tight integration of continuous features in TiMBL.} We
additionally employ the feature selection method of
\newcite{Baldwin:Bond:2003b}, which generates a combined ranking of all
features in descending order of ``informativeness'' and skims off the
top-$N$ features for use in classification; $N$ was set to 100 in all
experiments.

As observed above, a significant number of lexemes in the ERG occur in
multiple lexical items. If we were to take all lexical type combinations
observed for a single lexeme, the total number of lexical
``super''-types would be 451, of which 284 are singleton classes. Based
on the sparseness of this data and also the findings of
\newcite{Baldwin:Bond:2003b} over a countability learning task, we
choose to carry out DLA via a suite of 110 binary classifiers, one for
each lexical type.

We deliberately avoid carrying out extensive feature engineering over a
given secondary LR, choosing instead to take a varied but simplistic set
of features which is parallelled as much as possible between LRs (see
\secref[s]{sec:morphology}--\ref{sec:ontology} for details). We
additionally tightly constrain the feature space to a maximum of 3,900
features, and a maximum of 50 feature instances for each feature type;
in each case, the 50 feature instances are selected by taking the
features with highest saturation (i.e.\ the highest ratio of non-zero
values) across the full lexicon.  This is in an attempt to make
evaluation across the different secondary LRs as equitable as possible,
and get a sense of the intrinsic potential of each secondary LR in DLA.
Each feature instance is further translated into two feature values: the
raw count of the feature instance for the target word in question, and
the relative occurrence of the feature instance over all target word
token instances.

One potential shortcoming of our classifier architecture is that a given
word can be negatively classified by all unit binary classifiers and
thus not assigned any lexical items. In this case, we fall back on
the majority-class lexical type for each word class the word has been
pre-identified as belonging to.

\section{Morphology-based Deep Lexical Acquisition}
\label{sec:morphology}

We first perform DLA based on the following morphological LRs: (1) word
lists, and (2) morphological lexicons with a description of derivational
word correspondences. Note that in evaluation, we
presuppose that we have access to word lemmas although in the first
instance, it would be equally possible to run the method over
non-lemmatised data.\footnote{Although this would inevitably lose
  lexical generalisations among the different word forms of a given
  lemma.}


\subsection{Character \ngram[s]}
\label{sec:char-ngram}

In line with our desire to produce DLA methods which can be deployed
over both low- and high-density languages, our first feature
representation takes a simple word list and converts each lexeme into a
character \ngram representation.\footnote{We also experimented with
  syllabification, but found the character \ngram[s] to produce superior
  results.} In the case of English, we generated all 1- to 6-grams for
each lexeme, and applied a series of filters to: (1) filter out all
\ngram[s] which occurred less than 3 times in the lexicon data; and (2)
filter out all \ngram[s] which occur with the same frequency as larger
\ngram[s] they are proper substrings of. We then select the 3,900 character
\ngram[s] with highest saturation across the lexicon data (see
\secref{sec:classifier}).

The character \ngram-based classifier is the simplest of all classifiers
employed in this research, and can be deployed on any language for which
we have a word list (ideally lemmatised). 

\subsection{Derviational morphology}
\label{sec:derivational}

The second morphology-based DLA method makes use of derivational
morphology and analysis of the process of word formation. As an example
of how derivational information could assist DLA, knowing that the noun
\example{achievement} is deverbal and incorporates the \example{-ment}
suffix is a strong predictor of it being optionally uncountable and
optionally selecting for a PP argument (i.e.\ being of lexical type
\lextype{n\_mass\_count\_ppof\_le}).

We generate derivational morphological features for a given lexeme by
determining its word cluster in CATVAR\footnote{In the case that the a
  given lemma is not in CATVAR, we attempt to dehyphenate and then
  deprefix the word to find a match, failing which we look for the
  lexeme of smallest edit distance.} \cite{Habash:Dorr:2003} and then
for each sister lexeme (i.e. lexeme occurring in the same cluster as the
original lexeme with the same word stem), determine if there is a series
of edit operations over suffixes and prefixes which maps the lexemes
onto one another. For each sister lexeme where such a correspondence is
found to exist, we output the nature of the character transformation and
the word classes of the lexemes involved. E.g., the sister lexemes for
\example{achievement\wcnoun} in CATVAR are \example{achieve\wcverb},
\example{achiever\wcnoun}, \example{achievable\wcadj} and
\example{achievability\wcnoun}; the mapping between
\example{achievement\wcnoun} and \example{achiever\wcnoun}, e.g., would
be analysed as:
\begin{center}
\begin{tabular}{lp{2cm}}
  \texttt{N $-$ment\$ $\rightarrow$ N $+$r\$} &
\end{tabular}
\end{center}
Each such transformation is treated as a single feature.

We exhaustively generate all such transformations for each lexeme, and
filter the feature space as for character \ngram[s] above.

Clearly, LRs which document derivational morphology are typically only
available for high-density languages. Also, it is worth bearing in mind
that derivational morphology exists in only a limited form for certain
language families, e.g.\ agglutinative languages.

\begin{table*}
  \centering
  \smaller
  \begin{tabular}{lllr}
    \hline
    \hline
    \\
    & \raisedtab{\tabheader{Feature type}} & \raisedtab{\tabheader{Positions/description}} & \raisedtab{\tabheader{Total}}\\
    \hline
    \hline
    \multicolumn{2}{l}{\tabheader{TAGGER}} & & \tabheader{39}\\
    & POS tag & $(-4,-3,-2,-1,0,1,2,3,4)$ & 9\\
    & Word & $(-4,-3,-2,-1,1,2,3,4)$ & 8\\
    & POS bi-tag &
    $(\,(-4,-1),(-4,0),(-3,-2),(-3,-1),(-3,0),(-2,-1),(-2,0),$ \\
    & & $(-1,0),(0,1),(0,2),(0,3),(0,4),(1,2),(1,3),(1,4),(2,3)\,)$ &
    16 \\
    & Bi-word &
    $((-3,-2),(-3,-1),(-2,-1),(1,2),(1,3),(2,3))$ & 6 \\

    \hline

    \multicolumn{3}{l}{\tabheader{CHUNKER}} & \tabheader{39}\\
    & \modifier[head] & Chunk heads when target word is modifier & 1\\
    & \modifier[chunk] & Chunk types when target word is modifier & 1\\
    & \modifiee[word] & Modifiers when target word is chunk head & 1\\
    & \modifiee[POS] & POS tag of modifiers when target word
    is chunk head & 1\\
    & \modifiee[word+POS] & Word $+$ POS tag of modifiers when target word
    is chunk head & 1\\
    & POS tag & $(-3,-2,-1,0,1,2,3)$ & 7\\
    & Word & $(-3,-2,-1,1,2,3)$ & 6\\
    & Chunk & $(-4,-3,-2,-1,0,1,2,3,4)$ & 9\\
    & Chunk head & $(-3,-2,-1,1,2,3)$ & 6\\
    & Bi-chunk & $((-2,-1),(-2,0),(-1,0),(0,1),(0,2),(1,2))$ & 6\\
    \hline

    \multicolumn{3}{l}{\tabheader{DEPENDENCY PARSER}} & \tabheader{39}\\
    & POS tag & $(-2,-1,0,1,2)$ & 5\\
    & Word & $(-2,-1,1,2)$ & 4\\
    & \conj[word] & Words the target word coordinates with & 1\\
    & \conj[POS] & POS of words the target word coordinates with & 1\\
    & Head & Head word when target word modifier in dependency relation
    ($\times$ 14)& 14 \\
    & Modifier & Modifier when target word head of dependency relation
    ($\times$ 14)& 14 \\
    \hline
  \end{tabular}
  \mycaption{Feature types used in syntax-based DLA for the different preprocessors }
  \label{tab:features}
\end{table*}

\section{Syntax-based Deep Lexical Acquisition}
\label{sec:corpus}

Syntax-based DLA takes a raw text corpus and preprocesses it with either
a tagger, chunker or dependency parser. It then extracts a set of 39
feature types based on analysis of the token occurrences of a given
lexeme, and filters over each feature type to produce a maximum of 50
feature instances of highest saturation (e.g.\ if the feature type is
the word immediately proceeding the target word, the feature instances
are the 50 words which proceed the most words in our lexicon).  The
feature signature associated with a word for a given preprocessor type
will thus have a maximum of 3,900 items ($39\times 50\times
2$).\footnote{Note that we will have less than 50 feature instances for
  some feature types, e.g.\ the POS tag of the target word, given that
  the combined size of the Penn POS tagset is 36 elements (not including
  punctuation).}

\subsection{Tagging}
\label{sec:tagger}

The first and most basic form of syntactic preprocessing is part-of-speech (POS)
tagging. For our purposes, we use a Penn treebank-style tagger
custom-built using fnTBL 1.0 \cite{Ngai:Florian:2001}, and further
lemmatise the output of the tagger using morph \cite{Minnen:2000}.

The feature types used with the tagger are detailed in
\tabref{tab:features}, where the position indices are relative to the
target word (e.g.\ the word at position $-2$ is two words to the left of
the target word, and the POS tag at position $0$ is the POS of the
target word). All features are relative to the POS tags and words in the
immediate context of each token occurrence of the target word.
``Bi-words'' are word bigrams (e.g.\ bi-word $(1,3)$ is the bigram made up
of the words one and three positions to the right of the target word);
``bi-tags'' are, similarly, POS tag bigrams.

\subsection{Chunking}
\label{sec:chunker}

The second form of syntactic preprocessing, which builds directly on the
output of the POS tagger, is CoNLL 2000-style full text chunking
\cite{TjongKimSang:Buchholz:2000}. The particular chunker we use was
custom-built using fnTBL 1.0 once again, and operates over the
lemmatised output of the POS tagger.

The feature set for the chunker output includes a subset of the POS
tagger features, but also makes use of the local syntactic structure in
the chunker input in incorporating both intra-chunk features (such as
modifiers of the target word if it is the head of a chunk, or the head
if it is a modifier) and inter-chunk features (such as surrounding chunk
types when the target word is chunk head). See \tabref{tab:features} for
full details. 

Note that while chunk parsers are theoretically easier to develop than
full phrase-structure or treebank parsers, only high-density languages
such as English and Japanese have publicly available chunk parsers.

\subsection{Dependency parsing}
\label{sec:parser}

The third and final form of syntactic preprocessing is dependency
parsing, which represents the pinnacle of both robust syntactic
sophistication and inaccessibility for any other than the
highest-density languages.

The particular dependency parser we use is RASP\footnote{RASP is,
  strictly speaking, a full syntactic parser, but we use it in
  dependency parser mode} \cite{Briscoe:Carroll:2002}, which outputs
head--modifier dependency tuples and further classifies each tuple
according to a total of 14 relations; RASP also outputs the POS tag of
each word token. As our features, we use both local word and POS
features, for comparability with the POS tagger and chunker, and also
dependency-derived features, namely the modifier of all dependency
tuples the target word occurs as head of, and conversely, the head of
all dependency tuples the target word occurs as modifier in, along with
the dependency relation in each case. See \tabref{tab:features} for full
details.

\subsection{Corpora}

We ran the three syntactic preprocessors over a total of three corpora,
of varying size: the Brown corpus ($\sim$460K tokens) and Wall Street
Journal corpus ($\sim$1.2M tokens), both derived from the Penn Treebank
\cite{Marcus:1993}, and the written component of the British National
Corpus ($\sim$98M tokens: \newcite{Burnard:2000}). This selection is
intended to model the effects of variation in corpus size, to
investigate how well we could expect syntax-based DLA methods to perform
over both smaller and larger corpora.

Note that the only corpus annotation we make use of is sentence
tokenisation, and that all preprocessors are run automatically over the
raw corpus data. This is in an attempt to make the methods maximally
applicable to lower-density languages where annotated corpora tend not
to exist but there is at least the possibility of accessing raw text
collections. 

\section{Ontology-based Deep Lexical Acquisition}
\label{sec:ontology}

The final DLA method we explore is based on the hypothesis that there is
a strong correlation between the semantic and syntactic similarity of
words, a claim which is best exemplified in the work of
\newcite{Levin:1993} on diathesis alternations.  In our case, we take
word similarity as given and learn the syntactic behaviour of novel
words relative to semantically-similar words for which we know the
lexical types. We use WordNet 2.0 \cite{Fellbaum:1998} to determine word
similarity, and for each sense of the target word in WordNet: (1)
construct the set of ``semantic neighbours'' of that word sense,
comprised of all synonyms, direct hyponyms and direct hypernyms; and (2)
take a majority vote across the lexical types of the semantic neighbours
which occur in the training data. Note that this diverges from the
learning paradigm adopted for the morphology- and syntax-based DLA
methods in that we use a simple voting strategy rather than relying on
an external learner to carry out the classification. The full set of
lexical entries for the target word is generated by taking the union of
the majority votes across all senses of the word, such that a polysemous
lexeme can potentially give rise to multiple lexical entries. This
learning procedure is based on the method used by
\newcite{VanderBeek:Baldwin:2004} to learn Dutch countability.

As for the suite of binary classifiers, we fall back on the majority
class lexical type as the default in the instance that a given lexeme is
not contained in WordNet 2.0 or no classification emerges from the set
of semantic neighbours. It is important to realise that WordNet-style
ontologies exist only for the highest-density languages, and that this
method will thus have very limited language applicability.

\begin{figure*}[p]
  \begin{center}
  \begin{minipage}[t]{0.47\linewidth}
    \epsfig{file=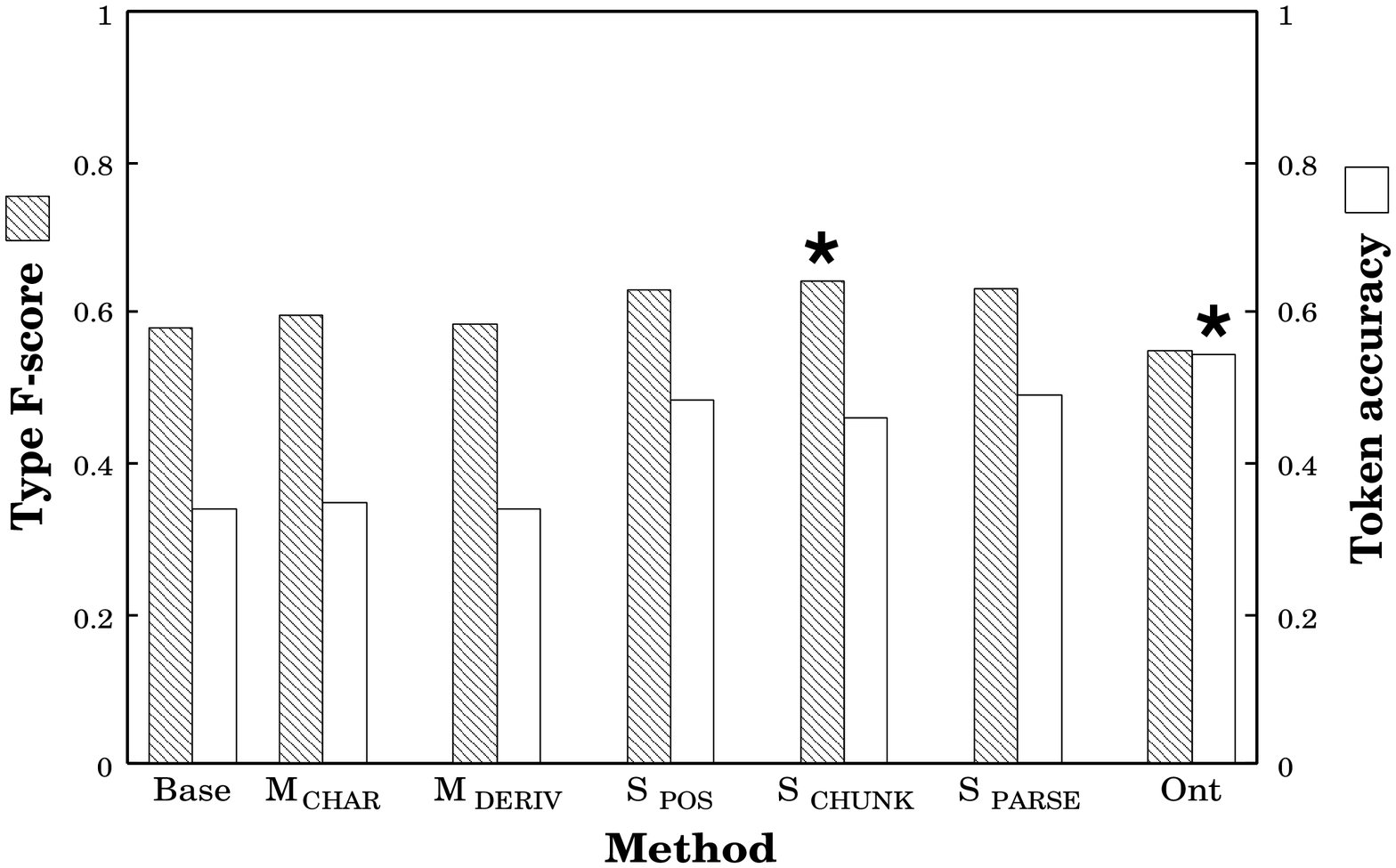,width=\textwidth}
    \mycaption{Results for the proposed deep lexical acquisition methods over
      \textsc{all} lexical types}
    \label{fig:results-all}

    \vspace*{5ex}

    \epsfig{file=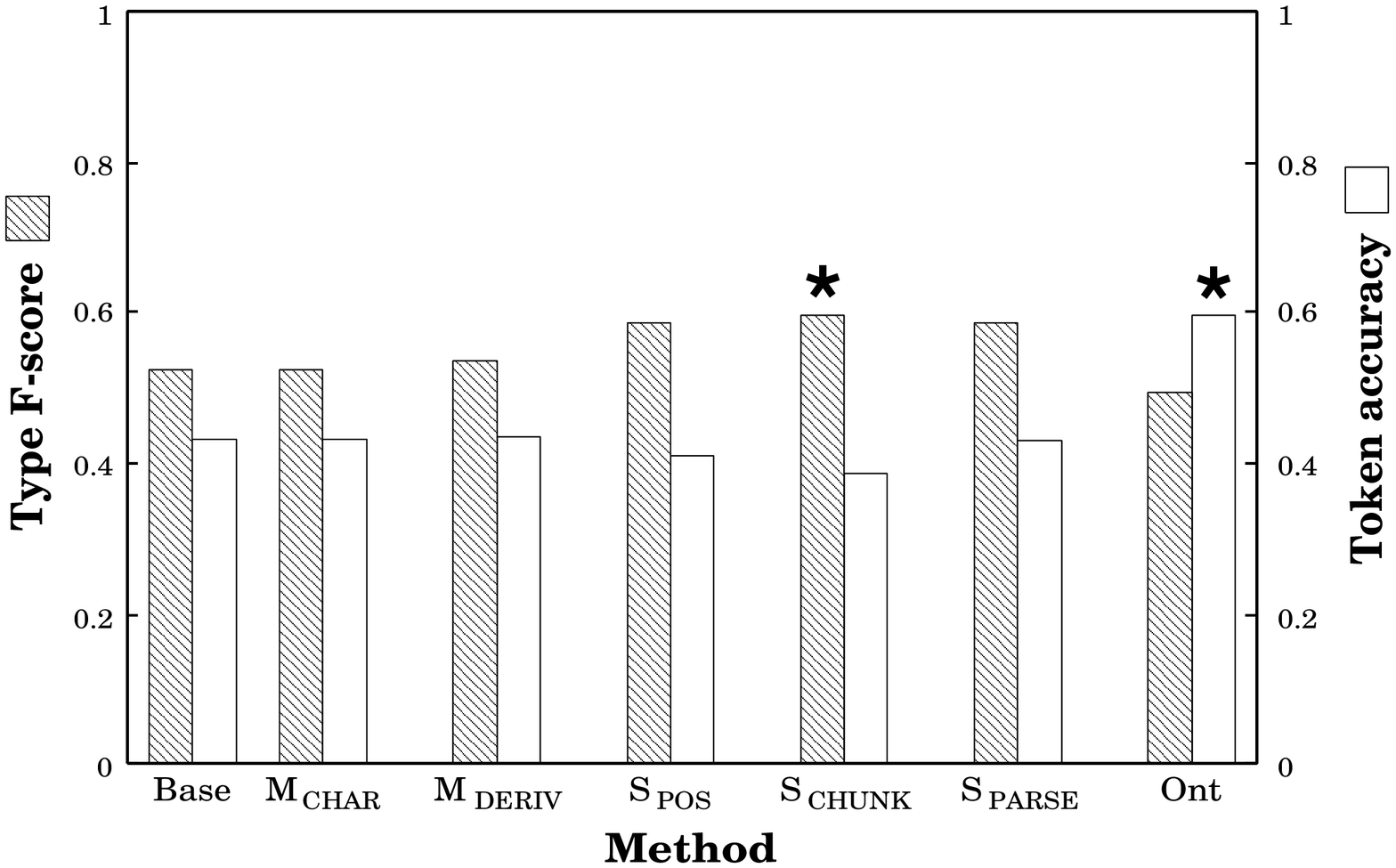,width=\textwidth}
    \mycaption{Results for the proposed deep lexical acquisition methods over
      \textsc{noun} lexical types}
    \label{fig:results-noun}

    \vspace*{5ex}

    \epsfig{file=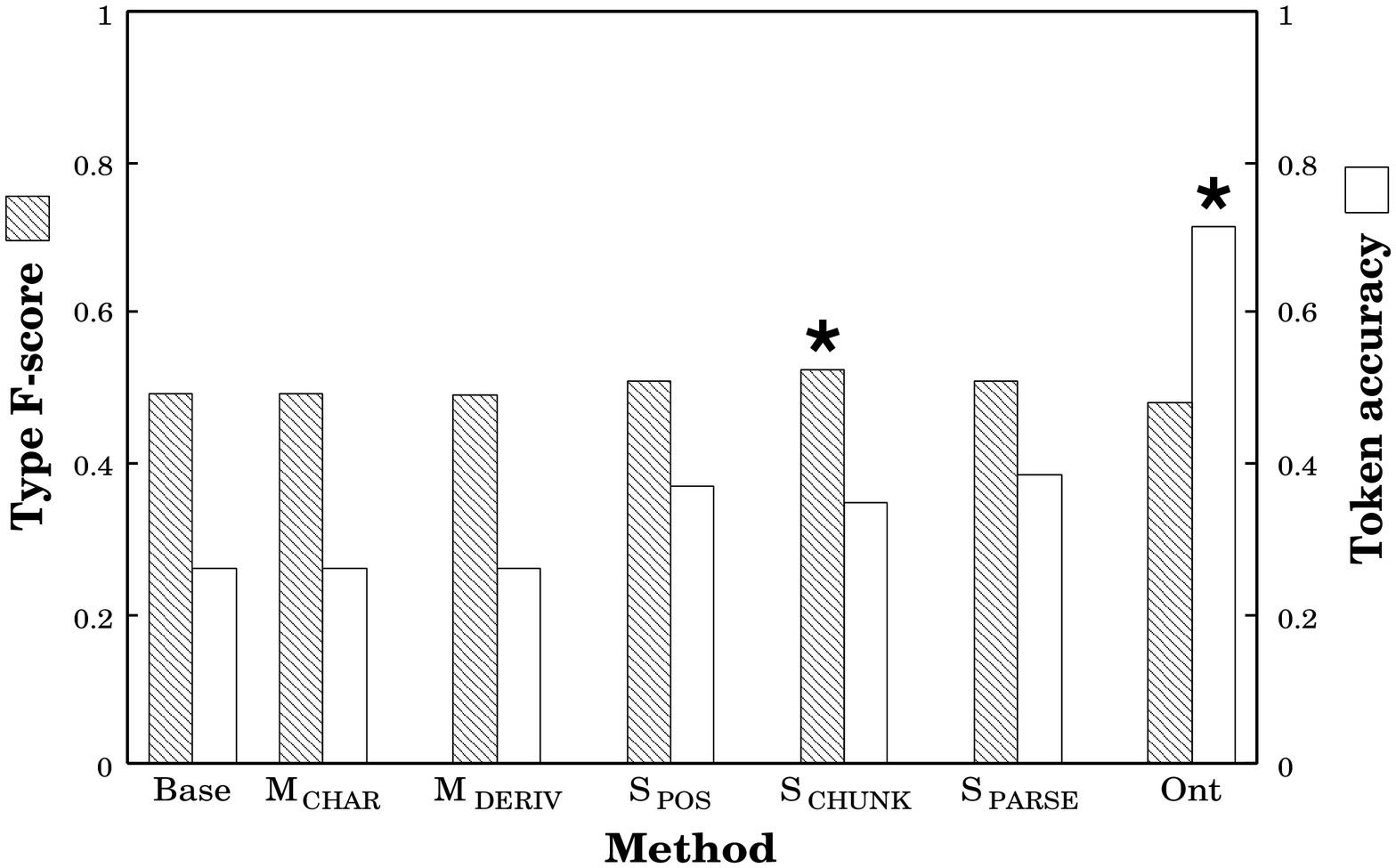,width=\textwidth}
    \mycaption{Results for the proposed deep lexical acquisition methods over
      \textsc{verb} lexical types}
    \label{fig:results-verb}
  \end{minipage}
  \hfill
  \begin{minipage}[t]{0.47\linewidth}
    \epsfig{file=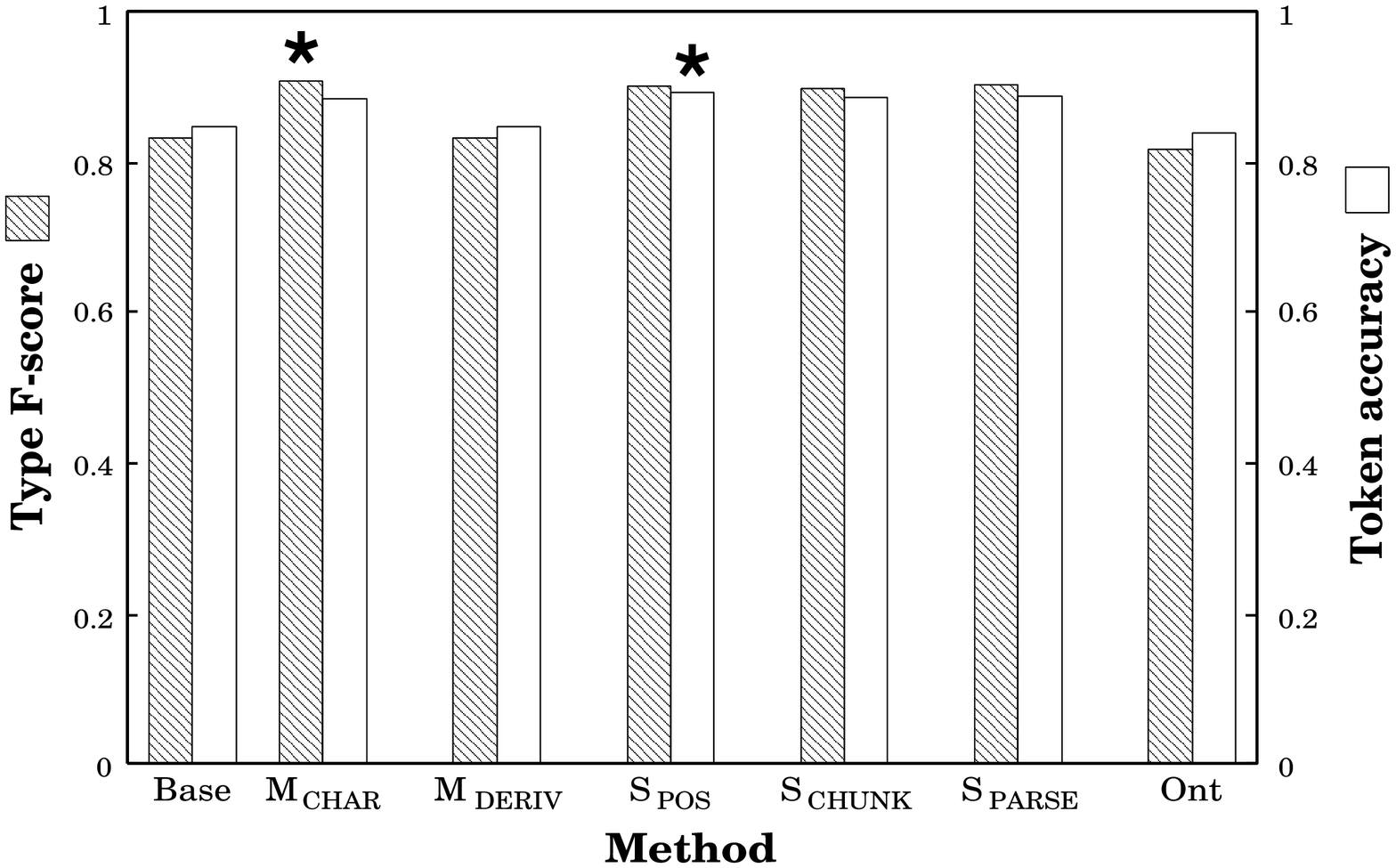,width=\textwidth}
    \mycaption{Results for the proposed deep lexical acquisition methods over
      \textsc{adjective} lexical types}
    \label{fig:results-adj}

    \vspace*{5ex}

    \epsfig{file=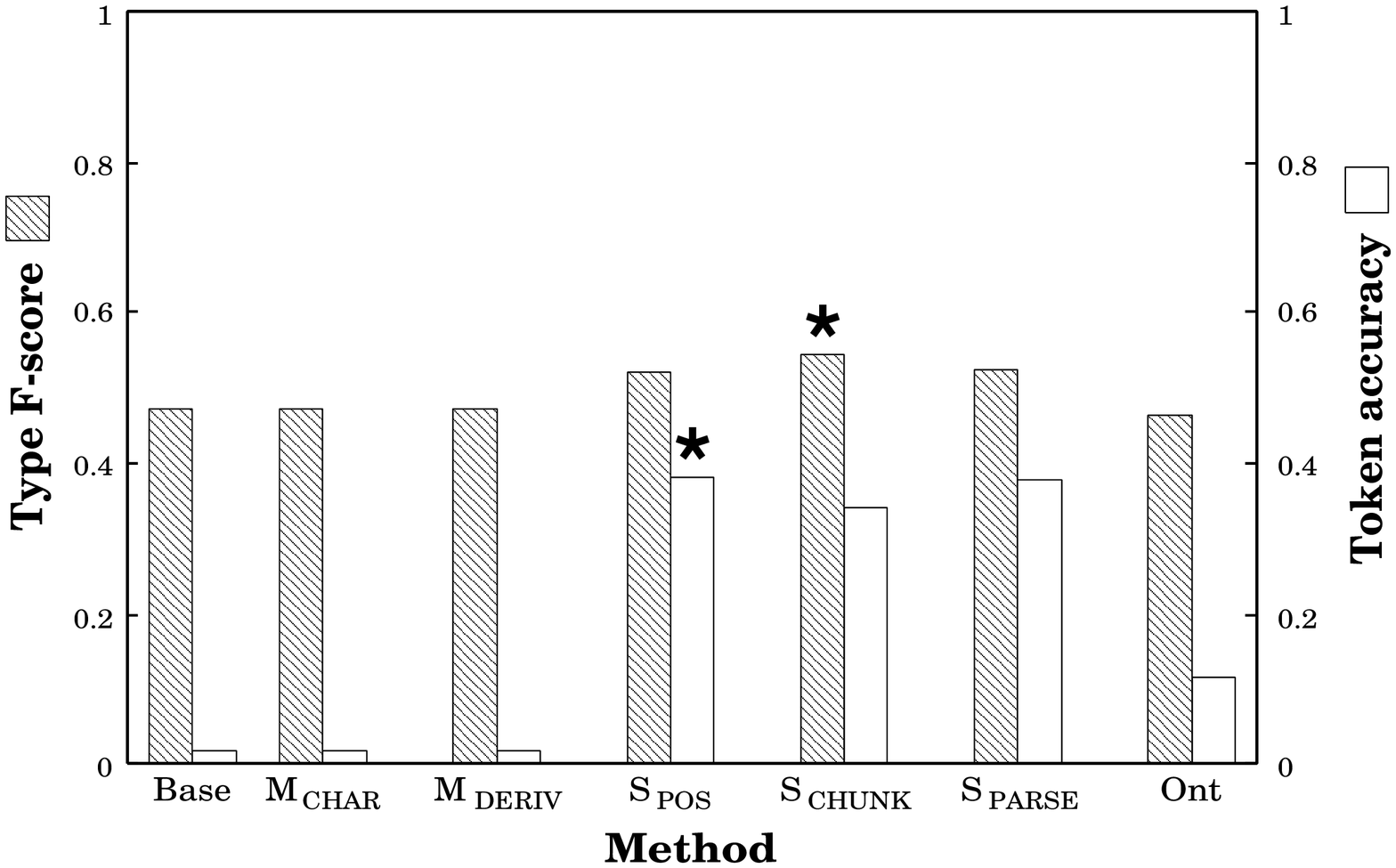,width=\textwidth}
    \mycaption{Results for the proposed deep lexical acquisition methods over
      \textsc{adverb} lexical types}
    \label{fig:results-adv}

    \vspace*{5ex}

    \epsfig{file=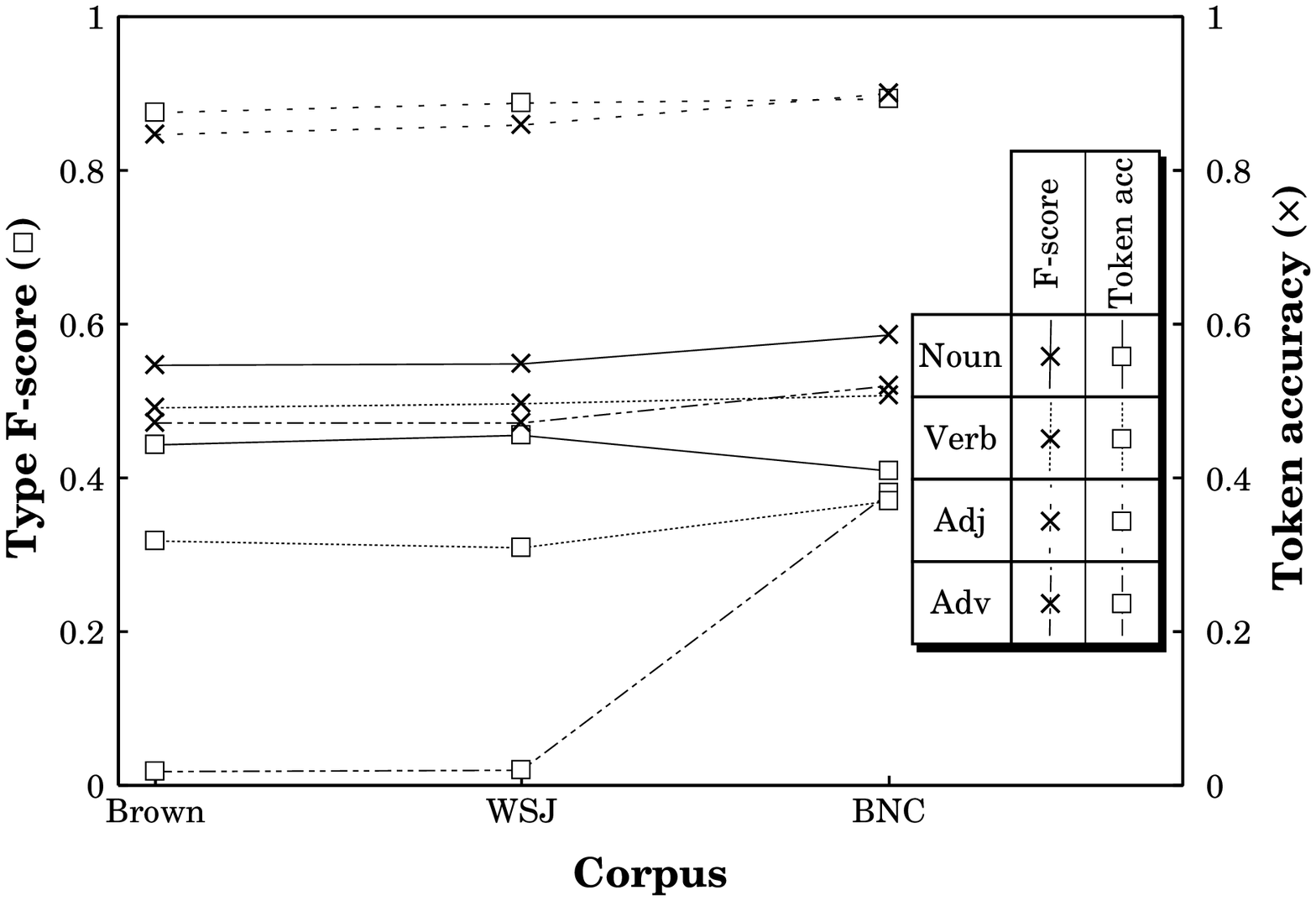,width=\textwidth}
    \mycaption{Results for the syntax-based deep lexical acquisition methods
      over corpora of differing size}
    \label{fig:results-token}

    \vspace*{4ex}

  \end{minipage}
  \end{center}
  \textit{Note:} {Base} = baseline, {M$_\mathrm{CHAR}$} =
  morphology-based DLA with character \ngram[s], {M$_\mathrm{DERIV}$} =
  derivational morphology-based DLA, {S$_\mathrm{POS}$} = syntax-based
  DLA with POS tagging, {S$_\mathrm{CHUNK}$} = syntax-based DLA with
  chunking, {S$_\mathrm{PARSE}$} = syntax-based DLA with dependency
  parsing, and {Ont} = ontology-based DLA
\end{figure*}

\section{Evaluation}
\label{sec:evaluation}

We evaluate the component methods over the 5,675 open-class lexical
items of the ERG described in \secref{sec:erg} using 10-fold stratified
cross-validation. In each case, we calculate the \tx{type precision}
(the proportion of correct hypothesised lexical entries) and \tx{type
  recall} (the proportion of gold-standard lexical entries for which we
get a correct hit), which we roll together into the \tx{type F-score}
(the harmonic mean of the two) relative to the gold-standard ERG
lexicon. We also measure the \tx{token accuracy} for the lexicon derived
from each method, relative to the Redwoods treebank of Verbmobil data
associated with the ERG (see \secref{sec:erg}).\footnote{Note that the
  token accuracy is calculated only over the open-class lexical items,
  not the full ERG lexicon.} The token accuracy represents a weighted
version of type precision, relative to the distribution of each lexical
item in a representative text sample, and provides a crude approximation
of the impact of each DLA method on parser coverage. That is, it gives
more credit for a method having correctly hypothesised a
commonly-occurring lexical item than a low-frequency lexical item, and
no credit for having correctly identified a lexical item not occurring
in the corpus.

The overall results are presented in \figref{fig:results-all}, which are
then broken down into the four open word classes in
\figref[s]{fig:results-noun}--\ref{fig:results-adv}. The baseline method
(\tablabel{Base}) in each case is a simple majority-class classifier,
which generates a unique lexical item for each lexeme pre-identified as
belonging to a given word class of the following type:
\begin{quote}
  \begin{tabular}{ll}
    Word class & Majority-class lexical type \\
    \hline
    Noun & \lextype{n\_intr\_le} \\
    Verb & \lextype{v\_np\_trans\_le} \\
    Adjective & \lextype{adj\_intrans\_le} \\
    Adverb & \lextype{adv\_int\_vp\_le} \\
  \end{tabular}
\end{quote}
In each graph, we present the type F-score and token accuracy for each
method, and mark the best-performing method in terms of each of these
evaluation measures with a star (\textbf{$\star$}). The results for
syntax-based DLA (\tablabel{S$_\mathrm{POS}$},
\tablabel{S$_\mathrm{CHUNK}$} and \tablabel{S$_\mathrm{PARSE}$}) are
based on the BNC in each case.  We return to investigate the impact of
corpus size on the performance of the syntax-based methods below.

Looking first at the combined results over all lexical types
(\figref{fig:results-all}), the most successful method in terms of type
F-score is syntax-based DLA, with chunker-based preprocessing marginally
outperforming tagger- and parser-based preprocessing (type F-score =
0.641). The most successful method in terms of token accuracy is
ontology-based DLA (token accuracy = 0.544).

The figures for token accuracy require some qualification:
ontology-based DLA tends to be liberal in its generation of lexical
items, giving rise to over 20\% more lexical items than the other
methods (7,307 vs. 5-6000 for the other methods) and proportionately low
type precision. This correlates with an inherent advantage in terms of token
accuracy, which we have no way of balancing up in our token-based
evaluation, as the treebank data offers no insight into the true worth
of false negative lexical items (i.e.\ have no way of distinguishing
between unobserved lexical items which are plain wrong from those which
are intuitively correct and could be expected to occur in alternate sets
of treebank data). We leave investigation of the impact of these extra
lexical items on the overall parser performance (in terms of chart
complexity and parse selection) as an item for future research.

The morphology-based DLA methods were around baseline performance
overall, with character \ngram[s] marginally more successful than
derivational morphology in terms of both type F-score and token
accuracy. 

Turning next to the results for the proposed methods over nouns, verbs,
adjectives and adverbs
(\figref[s]{fig:results-noun}--\ref{fig:results-adv}, respectively), we
observe some interesting effects. First, morphology-based DLA hovers around
baseline performance for all word classes except adjectives, where
character \ngram[s] produce the highest F-score of all methods, and
nouns, where derivational morphology seems to aid DLA slightly
(providing weak support for our original hypothesis in
\secref{sec:derivational} relating to deverbal nouns and affixation).

Syntax-based DLA leads to the highest type F-score for nouns, verbs and
adverbs, and the highest token accuracy for adjectives and adverbs. The
differential in results between syntax-based DLA and the other methods
is particularly striking for adverbs, with a maximum type F-score of
0.544 (for chunker-based preprocessing) and token accuracy of 0.340 (for
tagger-based preprocessing), as compared to baseline figures of 0.471
and 0.017 respectively. There is relatively little separating the three
styles of preprocessing in syntax-based DLA, although chunker-based
preprocessing tends to have a slight edge in terms of type F-score, and
tagger-based preprocessing generally produces the highest token
accuracy.\footnote{This trend was observed across all three corpora,
  although we do no present the full results here.} This suggests that
access to a POS tagger for a given language is sufficient to make
syntax-based DLA work, and that syntax-based DLA thus has moderately
high applicability across languages of different densities.

Ontology-based DLA is below baseline in terms of type F-score for all
word classes, but results in the highest token accuracy of all methods
for nouns and verbs (although this finding must be taken with a grain of
salt, as noted above).

Another noteworthy feature of
\figref[s]{fig:results-noun}--\ref{fig:results-adv} is the huge
variation in absolute performance across the word classes: adjectives
are very predictable, with a majority class-based baseline type F-score
of 0.832 and token accuracy of 0.847; adverbs, on the other hand, are
similar to verbs and nouns in terms of their baseline type F-score (at
0.471), but the adverbs that occur commonly in corpus data appear to
belong to less-populated lexical types (as seen in the baseline token
accuracy of a miniscule 0.017). Nouns appear the hardest to learn in
terms of the relative increment in token accuracy over the baseline.
Verbs are extremely difficult to get right at the type level, but
it appears that ontology-based DLA is highly adept at getting the
commonly-occurring lexical items right.

To summarise these findings, adverbs seem to benefit the most from
syntax-based DLA. Adjectives, on the other hand, can be learned most
effectively from simple character \ngram[s], i.e.\ similarly-spelled
adjectives tend to have similar syntax, a somewhat surprising finding.
Nouns are surprisingly hard to learn, but seem to benefit to some degree
from corpus data and also ontological similarity.  Lastly, verbs pose a
challenge to all methods at the type level, but ontology-based DLA seems
to be able to correctly predict the commonly-occurring lexical entries.


Finally, we examine the impact of corpus size on the performance of
syntax-based DLA with tagger-based preprocessing.\footnote{The results
  for chunker- and parser-based preprocessing are almost identical, and
  this omitted from the paper.}  In \figref{fig:results-token}, we
examine the relative change in type F-score and token accuracy across
the four word classes as we increase the corpus size (from 0.5m words to
1m and finally 100m words, in the form of the Brown corpus, WSJ corpus and BNC,
respectively). For verbs and adjectives, there is almost no change in
either type F-score or token accuracy when we increase the corpus size,
whereas for nouns, the token accuracy actually drops slightly. For
adverbs, on the other hand, the token accuracy jumps up from 0.020 to
0.381 when we increase the corpus size from 1m words to 100m words,
while the type F-score rises only slightly. It thus seems to be the case
that large corpora have a considerable impact on DLA for
commonly-occurring adverbs, but that for the remaining word classes, it
makes little difference whether we have 0.5m or 100m words. This can be
interpreted either as evidence that modestly-sized corpora are good
enough to perform syntax-based DLA over (which would be excellent news
for low-density languages!), or alternatively that for the simplistic
syntax-based DLA methods proposed here, more corpus data is not the solution
to achieving higher performance.

Returning to our original question of the ``bang for the buck''
associated with individual LRs, there seems to be no simple answer:
simple word lists are useful in learning the syntax of adjectives in
particular, but offer little in terms of learning the other three word
classes. Morphological lexicons with derivational information are
moderately advantageous in learning the syntax of nouns but little else.
A POS tagger seems sufficient to carry out syntax-based DLA, and the
word class which benefits the most from larger amounts of corpus data is
adverbs, otherwise the proposed syntax-based DLA methods don't seem to
benefit from larger-sized corpora. Ontologies have the greatest impact
on verbs and, to a lesser degree, nouns. Ultimately, this seems to lend
weight to a ``horses for courses'', or perhaps ``resources for courses''
approach to DLA.

\section{Conclusion}
\label{sec:discussion}

We have proposed three basic paradigms for deep lexical acquisition,
based on morphological, syntactic and ontological language resources,
and demonstrated the effectiveness of each strategy at learning lexical
items for the lexicon of a precision English grammar. We discovered
surprising variation in the results for the different DLA methods, with
each learning method performing particularly well for at least one basic
word class, but the best overall methods being syntax- and
ontology-based DLA. 

The results presented in this paper are based on one particular language
(English) and a very specific style of DLR (a precision grammar, namely
the English Resource Grammar), so some caution must be exercised in
extrapolating the results too liberally over new languages/DLA tasks.
In future research, we are interested in carrying out experiments over
other languages and alternate DLRs to determine how well these
results generalise and formulate alternate strategies for DLA.

\subsection*{Acknowledgements}
\setlength{\baselineskip}{0.8\baselineskip} 

{\small This material is based upon work supported in part by NTT
  Communication Science Laboratories, Nippon Telegraph and Telephone
  Corporation. We would like to thank the members of the University of
  Melbourne LT group and the three anonymous reviewers for their
  valuable input on this research.  }

\small  
\bibliographystyle{acl}

\end{document}